\documentclass[letterpaper, 10 pt, conference]{ieeeconf}  

\IEEEoverridecommandlockouts                              

\usepackage{graphicx}
\usepackage{color}
\usepackage{subfig}
\usepackage{mathtools}
\usepackage{rotating}    
\usepackage{multirow}
\usepackage{amssymb}  
\usepackage[table]{xcolor}
\usepackage[table]{xcolor}    
\title{\LARGE \bf Touch attention Bayesian models for robotic\\ active haptic exploration of heterogeneous surfaces}

\author{ Ricardo Martins, Jo\~{a}o Filipe Ferreira, \textit{Member, IEEE}, Jorge Dias, \textit{Senior Member, IEEE}
\thanks{The research leading to these results has been partially supported by the Portuguese Foundation for Science and Technology (FCT) with scholarships for Ricardo Martins. Ricardo Martins, Jo\~{a}o Filipe Ferreira and Jorge Dias are with Institute of Systems and Robotics, Department of Electrical Engineering and Computers, University of Coimbra, 3030 Coimbra, Portugal. Jorge Dias is also with the Robotics Institute, Khalifa University, Abu Dhabi, UAE. {\tt\small \{rmartins, jfilipe, jorge\}@isr.uc.pt}.}}

\begin{document}

\maketitle
\thispagestyle{empty}
\pagestyle{empty}

\begin{abstract}
This work contributes to the development of active haptic exploration strategies of surfaces using robotic hands in environments with an unknown structure. The architecture of the proposed approach consists two main Bayesian models, implementing the touch attention mechanisms of the system. The model $\pi_{per}$ perceives and discriminates different categories of materials (haptic stimulus) integrating compliance and texture features extracted from haptic sensory data. The model $\pi_{tar}$ actively infers the next region of the workspace that should be explored by the robotic system, integrating the task information, the permanently updated saliency and uncertainty maps extracted from the perceived haptic stimulus map, as well as, inhibition-of-return mechanisms.

The experimental results demonstrate that the Bayesian model $\pi_{per}$ can be used to discriminate 10 different classes of materials with an average recognition rate higher than $90\%$. The generalization capability of the proposed models was demonstrated experimentally. The \textit{ATLAS} robot, in the simulation, was able to perform the following of a discontinuity between two regions made of different materials with a divergence smaller than $1cm$ (30 trials). The tests were performed in scenarios with 3 different configurations of the discontinuity. The Bayesian models have demonstrated the capability to manage the uncertainty about the structure of the surfaces and sensory noise to make correct motor decisions from haptic percepts.
\end{abstract}

\section{Introduction}
\label{sec:introduction}
The diversity of the sensory and actuation apparatus of the new generation of robotic systems \cite{Dietsch2010} provides the support required to introduce these platforms in complex and dynamic environments (eg: domestic tasks, healthcare services, entertainment). To deal with the high diversity of environmental noisy sensory signals and uncertainties associated with the structure of these scenarios, those robotic platforms are endowed with active perception and action systems arranged in action-perception loop architectures \cite{Ernst2004}. The integration of attention mechanisms in those architectures contributes to the increasing of the efficiency of the action-perception loop process. Attention mechanisms integrate the saliency of the different sensory stimulus according to the task objectives and assist the decision making. In robotics, attention mechanisms have been predominantly applied to vision and audition sensory domains \cite{ferreira2014b}.

\begin{figure}[h]
  \centering
  \includegraphics[width=0.30\textwidth]{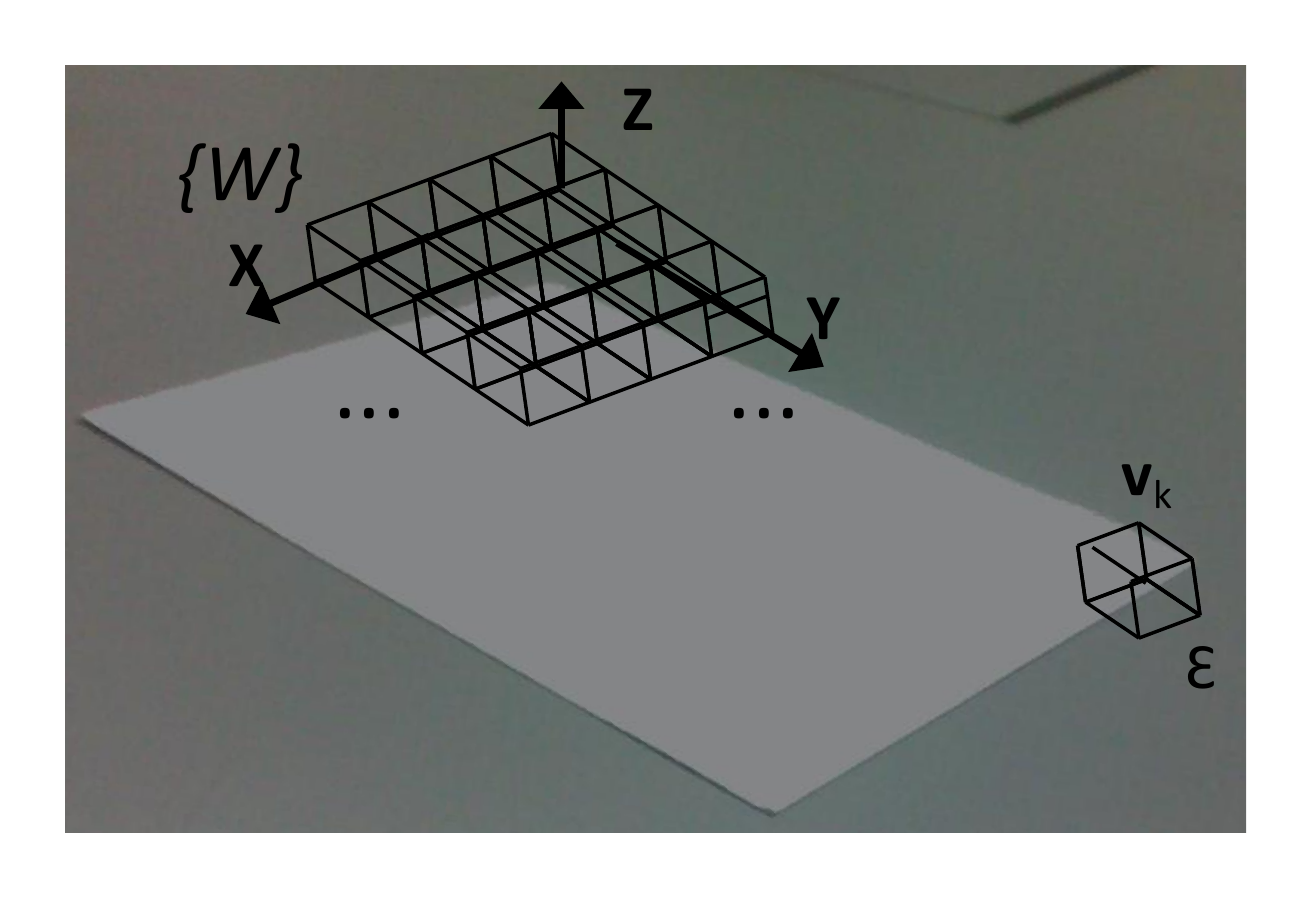}
  \caption{Partial representation of the volumetric grid framework integrated in a workspace region.}
  \label{fig:workspace_general_grid}
\end{figure}

Robotic haptic exploration \cite{Yousef2011} integrates haptic sensorial inputs (force, torque, tactile and temperature sensing) to perform tasks which are mainly involved in environments with low visibility conditions or partially occluded (underwater robotic manipulation, smoky and foggy disaster environments), in  service robotic platforms without vision systems or to complement vision systems information (eg: find, follow and extract the contour and structure of a napkin in the top of a table to subsequent grasping task).

This work proposes a formulation of Bayesian models implementing touch attention mechanisms involved in the active haptic exploration of unknown surfaces by generic robotic hands and sensory apparatus. The definition of the architecture of the Bayesian models, haptic sensory data processing pipeline, follows the principles described in the human anatomo-physiology studies \cite{Muller2008} \cite{Wacker2011} and on how humans manage uncertainty \cite{Ernst2004} to make motor decisions from percepts \cite{Faria2011}.

\begin{figure*}
  \centering
  \includegraphics[width=0.90\textwidth]{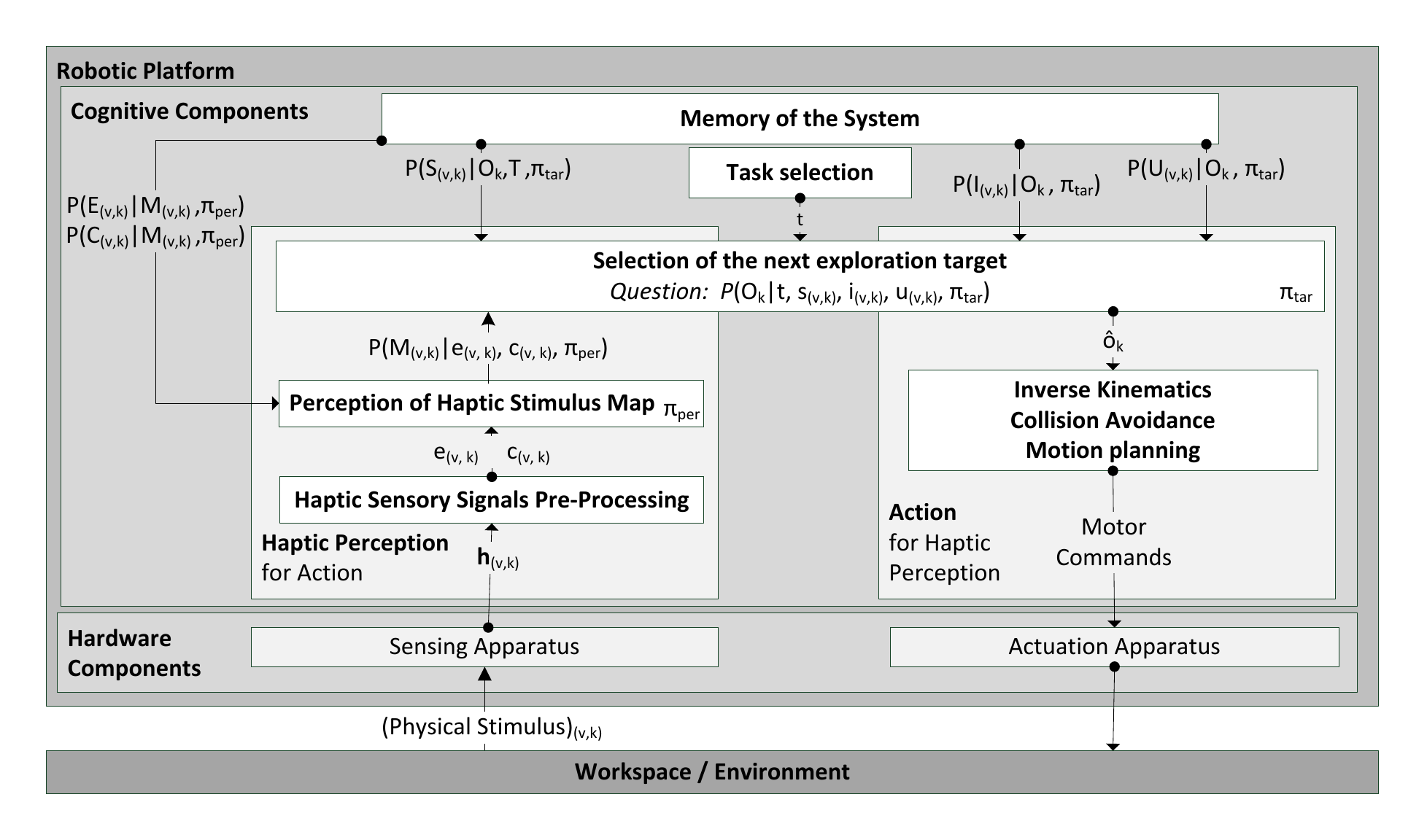}
  \caption{Global architecture of the system presented in this work. The variables of the system are summarized in table \ref{tb:main_variables}.}
  \label{fig:Architecture_and_information_flow_v1_5}
\end{figure*}

\section{Related works}
\label{ch:related_works}
The recent developments and improvements \cite{Lucarotti2013} \cite{Yousef2011} verified in the haptic sensing technologies (tactile, force, temperature) have promoted the intensive integration of these types of sensing technologies in the new generation of dexterous robotic hands, as presented in \cite{Dahiya2013}.  Due to the diversity of technologies and application fields, very distinct approaches and objectives have been followed to perform the robotic exploration of surfaces using haptic sensory inputs.

Some approaches perform the haptic exploration of surfaces with the objective of achieving a categorization of the surfaces or objects. The exploration is performed locally, assuming that the explored object is homogeneous or uniform in terms of the haptic features under analysis, such as the surface curvature \cite{Okamura2001}, texture \cite{oddo2011} \cite{Xu2013} \cite{Fishel2012} \cite{Chathuranga2013}, compliance \cite{martins2012b} \cite{Xu2013}, stickiness \cite{Liu2012} and thermal conductivity \cite{Xu2013} \cite{Castelli2008}. This work contributes to this group of approaches by  proposing a Bayesian model to discriminate 10 categories of materials  integrating compliance and texture features extracted from sensory data (from the work \cite{Xu2013}) corrupted with noise .

A second class of works integrates sensing, perception and local exploration mechanisms similar to the previous works, however they expand the exploration strategy to large and heterogeneous surfaces in the haptic features domain under analysis. The perceptual haptic map of the surface can be constructed following different strategies: the global exploration path is fixed and defined \textit{a-priori} \cite{Bologna2013} \cite{liu2010rolling} \cite{Barron2013}, the exploration is performed actively showing a active behaviour \cite{Martinez2013}, \cite{Li2013}. This work contributes to this class of approaches by proposing a formulation of Bayesian models implementing touch attention mechanisms involved in the active haptic exploration of unknown surfaces. Once this work assumes that the workspace is unknown \textit{a-priori to the system}, the exploration path is adapted actively by the touch attention mechanisms. This implementation strategy provides to the system the ability to deal with ambiguous sensory signals corrupted with noise and perform the active haptic exploration of surfaces with different geometries.

\section{Problem formulation and approach overview}
\label{ch:approach}
The main objective of a haptic exploration task consists in the determination of the sequence of states that the robotic system should follow to fulfil the objectives of the task. In the formulation of this work we assume that the exploration task is performed by a generic robotic system. Thus, the solution to the haptic exploration task is described in the tri-dimensional Cartesian space, by progressively determining the sequence of regions of the workspace that should be visited by the robotic platform during the task execution.

\begin{table}
\scriptsize
\centering
\caption{Summary of the relevant variables}
\begin{tabular}{|c|l|}
\hline
$v$ & Voxel of the workspace grid.\\ \hline
$k$ & Time / exploration iteration.\\ \hline
$M_{(v, k)}$ & Material category of $v$\\ \hline
$E_{(v, k)}$ & Texture characterization of $v$.\\ \hline
$C_{(v, k)}$ & Compliance characterization of $v$.\\ \hline
$\textbf{h}_{(v, k)}$ & Raw haptic sensing data acquired on $v$.\\ \hline
$O_k$ & Next workspace region to be explored. \\ \hline
$I_{(v,k)}$ & Inhibition level for voxel $v$.\\ \hline
$U_{(v,k)}$ & Uncertainty level for voxel $v$.\\ \hline
$S_{(v, k)}$ & Saliency of the perceived haptic stimulus in region $v$. \\ \hline
$T$ & Objective of the haptic exploration task. \\ \hline
\end{tabular}
\label{tb:main_variables}
\end{table}

The workspace where the robot operates can be associated with an inertial reference frame $\{\mathcal{W}\}$ and  delimited by the dimensions ${X}^W_{l} \leq x \leq {X}^W_{u}$, $ {Y}^W_{l} \leq y \leq {Y}^W_{u}$, ${Z}^W_{l} \leq z \leq {Z}^W_{u}$ that are the lower  and upper limits of the $X$, $Y$, $Z$ dimensions, respectively. In this work, the workspace of the robotic platform is partitioned in an isometric 3D grid (cubic voxels), as represented in figure \ref{fig:workspace_general_grid}. Each elementary cubic voxel $\textbf{v}_{k}$ has a side of dimension $\varepsilon$, is described by a 3D Cartesian location $(x, y, z)$ expressed in the inertial world referential $\{\mathcal{W}\}$  and can be associated to a random variable (inference grid), as will be presented throughout this manuscript.

Although the internal structure and configuration of the haptic stimulus disposed in the workspace is unknown \textit{a-priori} to the robotic system, the ground truth describing the target locations of the workspace that should be visited by the robotic platform during the task execution, can be formulated by an human operator for benchmark purposes and represented by $\mathcal{B} = \{\textbf{b}_{1}, \textbf{b}_{2}, \textbf{b}_{3}, \ldots,\textbf{b}_{k} \} $, $\textbf{b}_{i} = (x, y, z) \in \mathcal{R}^{3}$. The set of workspace regions visited by the robotic platform during the task execution can be represented by $\mathcal{V} = \{\textbf{v}_{1}, \textbf{v}_{2}, \textbf{v}_{3}, \ldots,\textbf{v}_{l} \}$, $\textbf{v}_{i} = (x, y, z) \in \mathcal{R}^{3}$.

The performance of the execution of the task by the robotic platform during an experimental trial can be evaluated by an error metric defined in equation \ref{eq:objective_definition}.

\begin{align}
\label{eq:objective_definition}
\Gamma= \sum_{i=1}^{l}\Vert \textbf{v}_{i}-\textbf{b}_{nearest}\Vert, \quad \text{given that} \nonumber \\
\forall_{\textbf{b}_{i}\in \mathcal{B}} \quad \exists_{\textbf{b}_{nearest}}:\Vert \textbf{v}_{i} - \textbf{b}_{nearest} \Vert \leq \Vert \textbf{v}_{i} - \textbf{b}_{i} \Vert 
\end{align}

Better autonomous exploration strategies provide lower values of $\Gamma$. This metric determines the total divergence between the exploration path executed by the robotic platform $\mathcal{V}$ and the benchmark path $\mathcal{B}$ defined by an external operator. The definition of this metric follows the same principles of analogous metrics proposed by \cite{Li2013} and \cite{Barron2013}.

\begin{figure}
  \centering
  \includegraphics[width=0.48\textwidth]{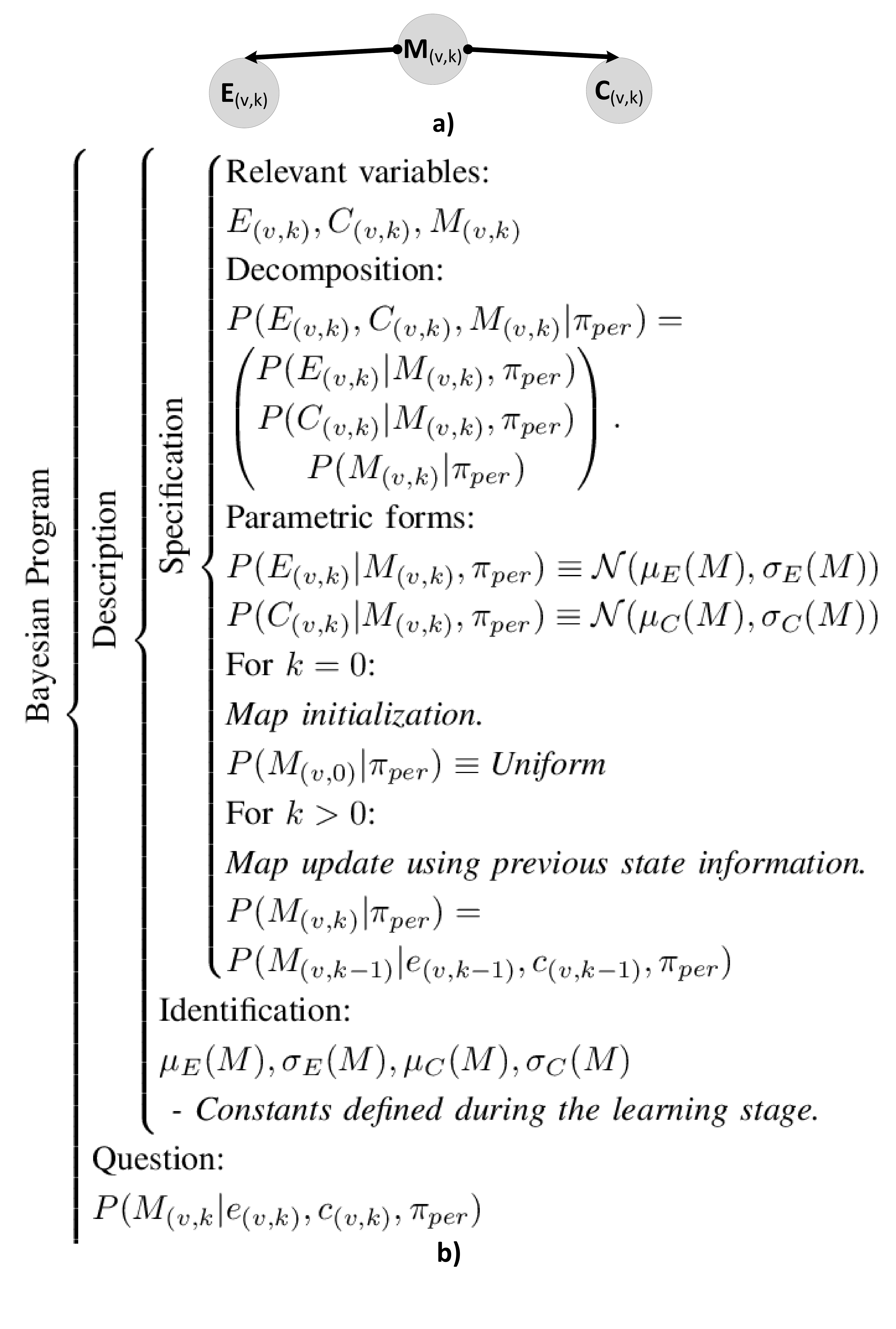}
  \caption{Bayesian model $\pi_{per}$:\textit{"Perception of haptic stimulus map"}. a) Graphical representation. b) Description of the Bayesian program.}
  \label{fig:Bayesian_Program_Network_Sub_v0}
\end{figure}

The global structure of the approach proposed in this work is presented in figure \ref{fig:Architecture_and_information_flow_v1_5}, arranged in a action-perception loop architecture. The haptic perception for action side of the loop integrates the Bayesian model $\pi_{per}$:\textit{"Perception of haptic stimulus map"}, while the action for haptic perception side of the loop integrates the Bayesian model $\pi_{tar}$:\textit{"Selection of the next exploration target"}. The main variables involved in this work are summarized in table \ref{tb:main_variables}.

\section{Perception of haptic stimulus map}
\label{ch:Bayesian_Program_I}

\subsection{Random variables of the model}
\label{ch:random_variables_of_the_model_perception}

Based on the sensory haptic inputs acquired at $v$, at each time iteration step $k$, the model $\pi_{per}$ implemented in  this Bayesian program determines the perceived category of material of the voxel $v$ of the workspace. This work considers that the robotic system has the capability to perceive and discriminate $n=10$ classes of different materials (haptic stimulus). The random variable $M_{(v, k)}=\text{"Material category of v"}$ can be defined as follows:

\begin{align}
M_{(v, k)} \in \{Material_1, \ldots, Material_{10}\}  
\end{align}

\textcolor{black}{We will consider the same set of materials that was used in the work \cite{Xu2013}. These 10 reference materials correspond to samples of acrylic, brick, copper, damp sponge, feather, rough foam, plush toy, silicone, soft foam, wood, respectively. These categories of materials are characterized by different properties of texture, compliance and thermal conductivity that were extracted using \textit{BioTac} biomimetic tactile sensor raw data (contact intensity, vibration, heat flow). In this work we will only consider the texture and compliance properties of the materials.}

The description of the texture and compliance properties of the region $v$ of the workspace is represented by the random variables $E_{(v, k)}= \text{"Texture characterization of v"}$ and $C_{(v, k)}= \text{"Compliance characterization of v"}$ respectively, $E_{(v, k)}=f(\textbf{h}_{(v, k)})$, $C_{(v, k)}=g(\textbf{h}_{(v, k)})$. The parameter $\textbf{h}_{(v, k)}$ represents haptic sensing measurements provided by the sensory apparatus of the robotic platform. The function $g$ transforms the haptic sensing measurements $\textbf{h}_{(v, k)}$ in a compliance characterization of the explored surface, while $f$ transforms  $\textbf{h}_{(v, k)}$ in a texture characterization of the surface. 

\subsection{Inference of the haptic stimulus category}
\label{ch:inference_of_the_haptic_stimulus_category}

The statistical independence relations between $E_{(v, k)}, C_{(v, k)}, M_{(v, k)}$ are expressed in figure \ref{fig:Bayesian_Program_Network_Sub_v0} a). Based on those statistical assumptions, the joint probability distribution function $P(E_{(v, k)}, C_{(v, k)}, M_{(v, k)},\pi_{per} )$ is decomposed as described in figure \ref{fig:Bayesian_Program_Network_Sub_v0} b). Each of those factors follows a probability distribution function presented in figure \ref{fig:Bayesian_Program_Network_Sub_v0} b). At each time iteration step, based on the observed data $e_{(v, k)}, c_{(v, k)}$, the Bayesian program described in figure \ref{fig:Bayesian_Program_Network_Sub_v0} b) is run with the question presented in equation \ref{eq:objectiveHapticStimulus}.

\begin{align}
P(M_{(v, k)}| e_{(v, k)}, c_{(v, k)},\pi_{per})=\nonumber \\ \frac{\begin{pmatrix}
P(e_{(v, k)}|M_{(v, k)},\pi_{per}).\\
P(c_{(v, k)}|M_{(v, k)},\pi_{per}).
P(M_{(v, k)},\pi_{per})
\end{pmatrix}}
{\sum \limits_{M_{(v, k)}}
\begin{pmatrix}
P(e_{(v, k)}|M_{(v, k)},\pi_{per}).\\
P(c_{(v, k)}|M_{(v, k)},\pi_{per}).
P(M_{(v, k)},\pi_{per})
\end{pmatrix}
}
\label{eq:objectiveHapticStimulus}
\end{align}

\subsection{Determination of $P(E_{(v, k)}|M_{(v, k)},\pi_{per})$ and $P(C_{(v, k)}|M_{(v, k)},\pi_{per})$ }
\label{sec:determination_of_probability_distribution_functions_perception}
\begin{figure}
  \centering
  \includegraphics[width=0.48\textwidth]{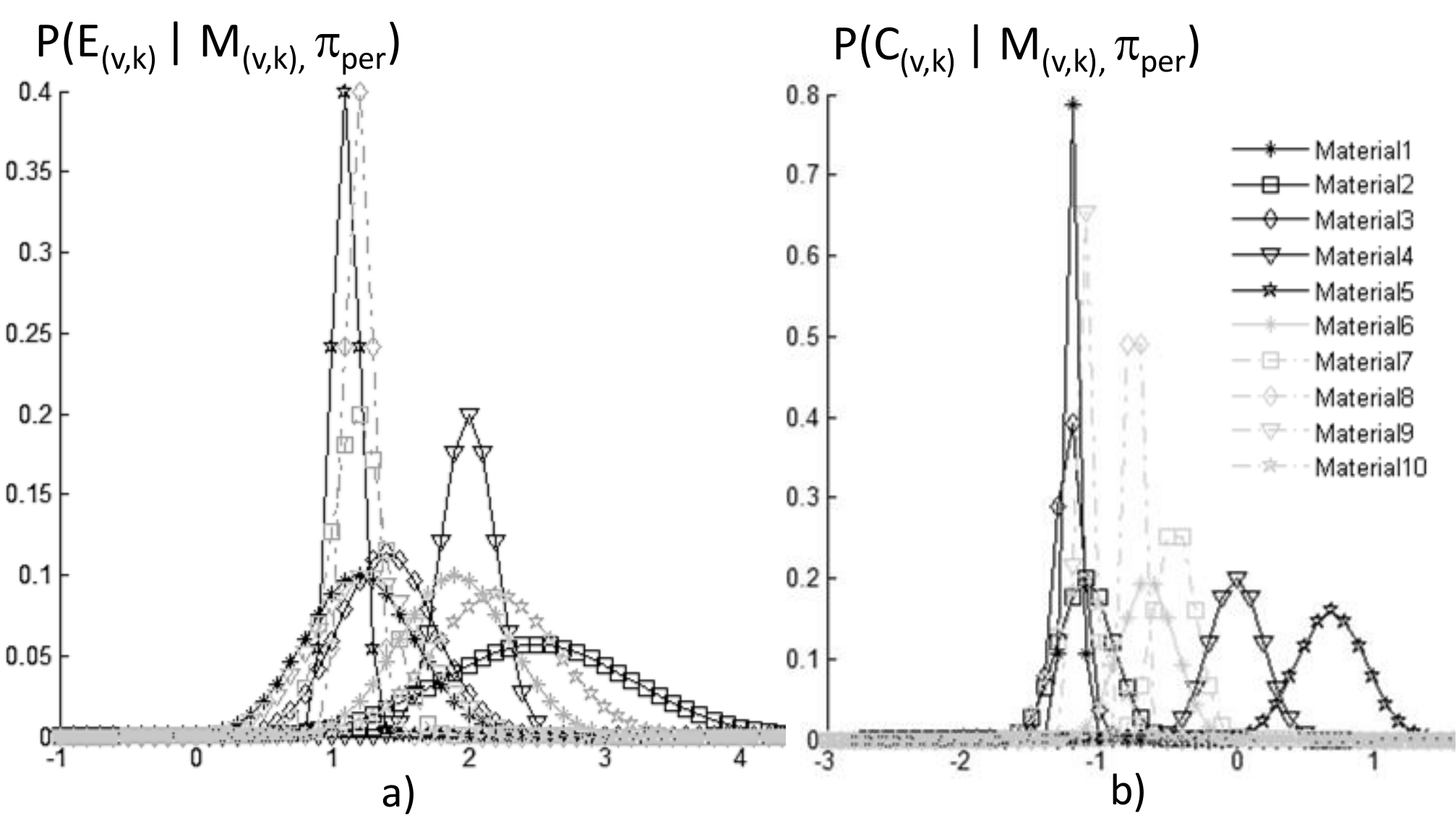}
  \caption{\textcolor{black}{Representation of} $P(E_{(vi, k)}|M_{(vi, k)},\pi_{per})$ (a)) and $P(C_{(vi, k)}|M_{(vi, k)},\pi_{per})$ \textcolor{black}{(b)) learned for 10 reference materials. Data extracted from} \cite{Xu2013}.}
  \label{fig:E-M_C-M}
\end{figure}

The free parameters $\mu_E(M)$, $\sigma_E(M)$, $\mu_C(M)$, $\sigma_C(M)$  of the Gaussian functions used to define the Normal probability distributions $P(E_{(v, k)} | M_{(v, k)},\pi_{per})$ and  $P(C_{(v, k)} | M_{(v, k)},\pi_{per})$ are estimated during experimental learning sessions. As described in \cite{Xu2013}, during the learning period, standard exploration procedures are performed for each of the $n=10$ reference materials. After the pre-determined number of standard explorations, the free parameters $\mu_E(M)$, $\sigma_E(M)$, $\mu_C(M)$, $\sigma_C(M)$ of the Normal ($\mathcal{N}$) distributions are determined by calculating the average and standard deviation of $E$ and $C$ for each reference material. The results are represented in the figures \ref{fig:E-M_C-M} a) and \ref{fig:E-M_C-M} b), extracting the data available from the work \cite{Xu2013}.

\section{Selection of the next exploration target}
\label{ch:motor_target_program}

\subsection{Random variables of the model}
\label{ch:random_variables_of_the_model_motor}
Based on the haptic stimulus $M_{(v,k)}$ map perceived during the exploration of $v$ ( previous section \ref{ch:Bayesian_Program_I}), at each time iteration step $k$, the model $\pi_{tar}$ implemented in this Bayesian program determines the next region of workspace that should be explored by the robotic system. This target is represented by the discrete random variable $O_k - \text{"Next workspace region to be explored"}$, given that $O_{k} \in \{v^1, v^2, v^3, \ldots, v^{\theta}\}$. $\theta$ is the total number of voxels in the grid representation of the workspace. $v^i$ is a compact representation of the voxel identifier.

The selection of $O_k$ is conditioned by inhibition-of-return mechanisms. The inhibition level imposed by inhibition-of-return process involved in the touch attention mechanisms is implemented by the continuous random variable $I_{(v,k)} - \text{"Inhibition level for voxel v."}$.

\begin{align}
\label{eq:inhibition_level_cell_v}
I_{(v,k)}=1-\Theta d^{\alpha-1}(1-d)^{1-\beta}, \quad  I_{(v,k)}\in [0, 1]  
\end{align}

In this work, due to the characteristics of the haptic exploration procedures presented in section \ref{sec:introduction}, the inhibition of return process promotes, at time iteration $k+1$, the exploration  of regions of the workspace different from the current position of the end-effector of the robotic system ($\hat{o}_{k-1}$). However, simultaneously, the inhibition-of-return process inhibits the exploration of regions too distant from $\hat{o}_{k-1}$, to avoid discontinuities in search and following exploration tasks. The inhibition levels $I_{(v,k)}$ for each voxel $v$ can be described by the equation \ref{eq:inhibition_level_cell_v}, considering $\alpha=1.01$ and $\beta=9$ (profile represented in figure   \ref{fig:probability_distribution_functions_v0} a)). The parameter $d$ is determined by $d=d_k / d_{max}$. The parameter $d_k$ expresses the Euclidean distance between $o_k$ and $\hat{o}_{k-1}$ and $d_{max}$ is a constant representing the maximum possible distance between $o_k$ and $\hat{o}_{k-1}$ for the workspace dimensions. $\Theta$ is a normalization constant. The values of $I_{(v,k)}(d)$ are ranged between $0$ and $1$. $I_{(v,k)}=0$ indicates that the inhibition-of-return mechanism applies no inhibition to voxel $v$, whereas $I_{(v,k)}=1$ indicates a full inhibition to voxel $v$.

\begin{figure}[h]
  \centering
  \includegraphics[width=0.48\textwidth]{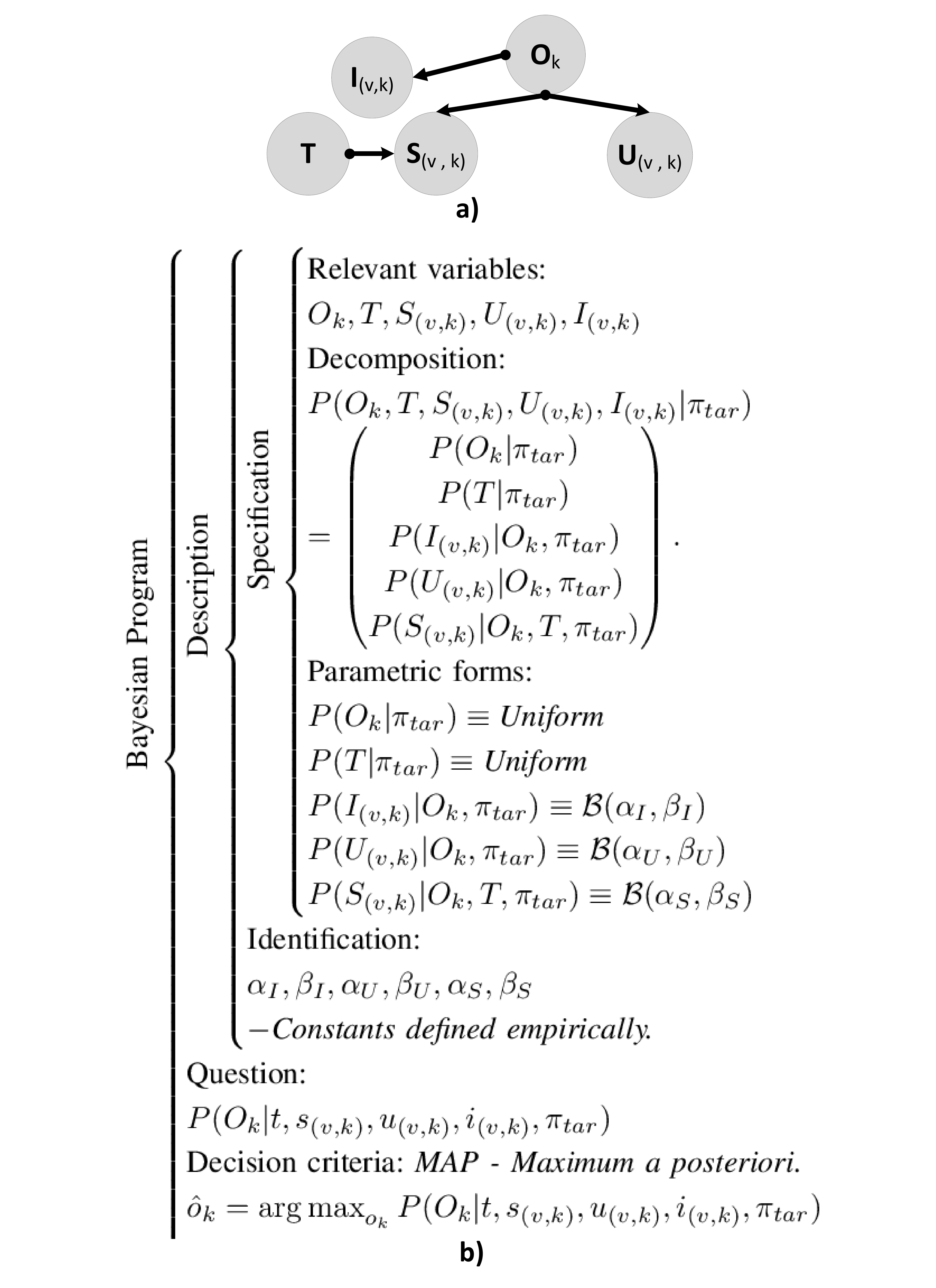}
  \caption{Bayesianl model $\pi_{tar}$:\textit{"Selection of the next exploration target"}. a) Graphical representation. b) Description of the Bayesian program.}
  \label{fig:Bayesian_Program_Network_Top_v0}
\end{figure}

The selection of the region $O_k$ of the workspace is also dependent of mechanisms to avoid the return to regions already explored and perceived with low uncertainty. In this work, those mechanisms are represented by the continuous random variable $U_{(v,k)}-\text{"Uncertainty level for voxel v."}$, described in equation \ref{eq:uncertainty_level_cell_v}. The operator $\mathcal{H}$ determines the entropy of the discrete random variable $M_{(v,k)}$. 

\begin{align}
\label{eq:uncertainty_level_cell_v}
U_{(v,k)}=\frac{\mathcal{H}(M_{(v,k)})}{max(\mathcal{H}(M_{(v,k)}))}, \quad U_{(v,k)} \in [0, 1]
\end{align}

Another factor conditioning the determination of $O_k$ is the saliency of the haptic stimulus perceived in the  region $v$ of the workspace and in its surroundings.  Besides depending on the perceived haptic stimulus $M_{(v,k)}$ map, the formulation of the saliency of those haptic stimulus is also dependent of the current objectives of the exploration task. The objectives of the task being executed by the robotic platform are represented by the discrete random variable $T=\text{"Task objective."}$, given that $T \in \{Task_1,\ldots ,Task_\Phi\}$. During an experimental trial the value of $T=t$ is considered constant in time $k$. $\Phi$ expresses the total number of tasks that can be executed by the robotic platform.

Based on these considerations, the saliency of the haptic stimulus perceived in the surroundings of $v$ can be formulated by the continuous random variable $S_{(v, k)}=\text{"Saliency of the perceived haptic stimulus in region v"}$. 
This work defines $S_{(v,k)}$ for  a class of tasks $T$=\textit{"Search and follow of discontinuities between regions of surfaces with $Material_a$ and $Material_{b}$."}, as presented in equation \ref{eq:ssDetermination}. $S_{(v,k)}$ is related by a soft evidence relation with the perceived haptic stimulus map $M_{(v,k)}$ .

\begin{align}
S=\frac{\max(|s_x|,|s_y|,|s_z|)}{s_{norm}}, \quad S_{(v, k)} \in [0,1]
\label{eq:ssDetermination}
\end{align}

The parameters $s_x=\mathcal{G}_{sobel_x}(\textbf{d})$, $s_y=\mathcal{G}_{sobel_y}(\textbf{d})$ and $s_z=\mathcal{G}_{sobel_z}(\textbf{d})$ are determined using the volumetric edge detector $\mathcal{\textbf{G}}_{sobel}$ following an approach analogous to the operator proposed in \cite{Bhattacharya96}. Considering that the exploratory element is located at $v$ of the workspace, a $26-th$ neighbourhood can be defined around that location. For a given neighbourhood $v0, \ldots, v26$, we can define $\textbf{d}=(\Omega_{(v0, k)}, \ldots, \Omega_{(v26, k)})$ as the set of values of $\Omega_{(vi, k)}$. We consider that the haptic stimulus perceived at each of the voxels $(v0,\ldots, v26)$ of a neighbourhood can be described by a probability distribution function $P(M_{(vi, k)}| e_{(vi, k)}, c_{(vi, k)},\pi_{per})$, respectively (details in section \ref{ch:Bayesian_Program_I}). We can define, for each region $vi$, a constant $\Omega_{(vi, k)}$ that expresses the similarity of the perceived material category of a region as $Material_a$ or $Material_b$. The constant $\Omega_{(vi, k)} \in [0,1] $ is determined by equation \ref{eq:omegaDetermination}.
 
\begin{align}
\Omega_{(vi, k)}=\frac{1-\begin{pmatrix}P(M_{(vi, k)}=Mat._b| e_{(vi, k)}, c_{(vi, k)},\pi_{per})-\\P(M_{(vi, k)}=Mat._a | e_{(vi, k)}, c_{(vi, k)},\pi_{per})\end{pmatrix}}{2}
\label{eq:omegaDetermination}
\end{align}

\subsection{Inference of the next exploration target}
\label{ch:inference_of_the_next_motor_target_location}

Based on the statistical independence relations between the random variables $O_k$, $I_{(v,k)}$, $U_{(v,k)}$, $S_{(v,k)}$, $T$, presented in figure \ref{fig:Bayesian_Program_Network_Top_v0} a), the probability joint distribution function $P(O_k, T,S_{(v,k)},U_{(v,k)}, I_{(v,k)} | \pi_{tar} )$ for this model $\pi_{tar}$, can be decomposed as summarized in figure \ref{fig:Bayesian_Program_Network_Top_v0} b). Each of those factors is described by a probability distribution function presented in figure \ref{fig:Bayesian_Program_Network_Top_v0} b). The final estimate for $\hat{o}_k$ is given via a \textit{Maximum a Posteriori} (MAP) decision rule, as expressed in equation \ref{eq:equationMain} given a specific task $T=t$.

\begin{align}
\hat{o}_k =\operatorname*{arg\,max}_{o_k} P(O_k | t, s_{(v,k)},i_{(v,k)}, u_{(v,k)}, \pi_{tar}) \nonumber\\
\hat{o}_k =\operatorname*{arg\,max}_{o_k} 
\begin{pmatrix}
P(t|\pi_{tar}).P(i_{(v,k)}|O_k,\pi_{tar}).\\
P(s_{(v,k)} | O_k, t,\pi_{tar}).P(u_{(v,k)} | O_k,\pi_{tar})
\end{pmatrix}\nonumber\\
\label{eq:equationMain}
\end{align}

\subsection{Determination of $P(S_{(v,k)}| O_k, T, \pi_{tar})$, $P(I_{(v,k)}|O_k,\pi_{tar})$, $P(U_{(v,k)}|O_k,\pi_{tar})$ }
\label{sec:determination_of_probability_distribution_functions_motor}

\begin{figure}[h]
  \centering
  \includegraphics[width=0.50\textwidth]{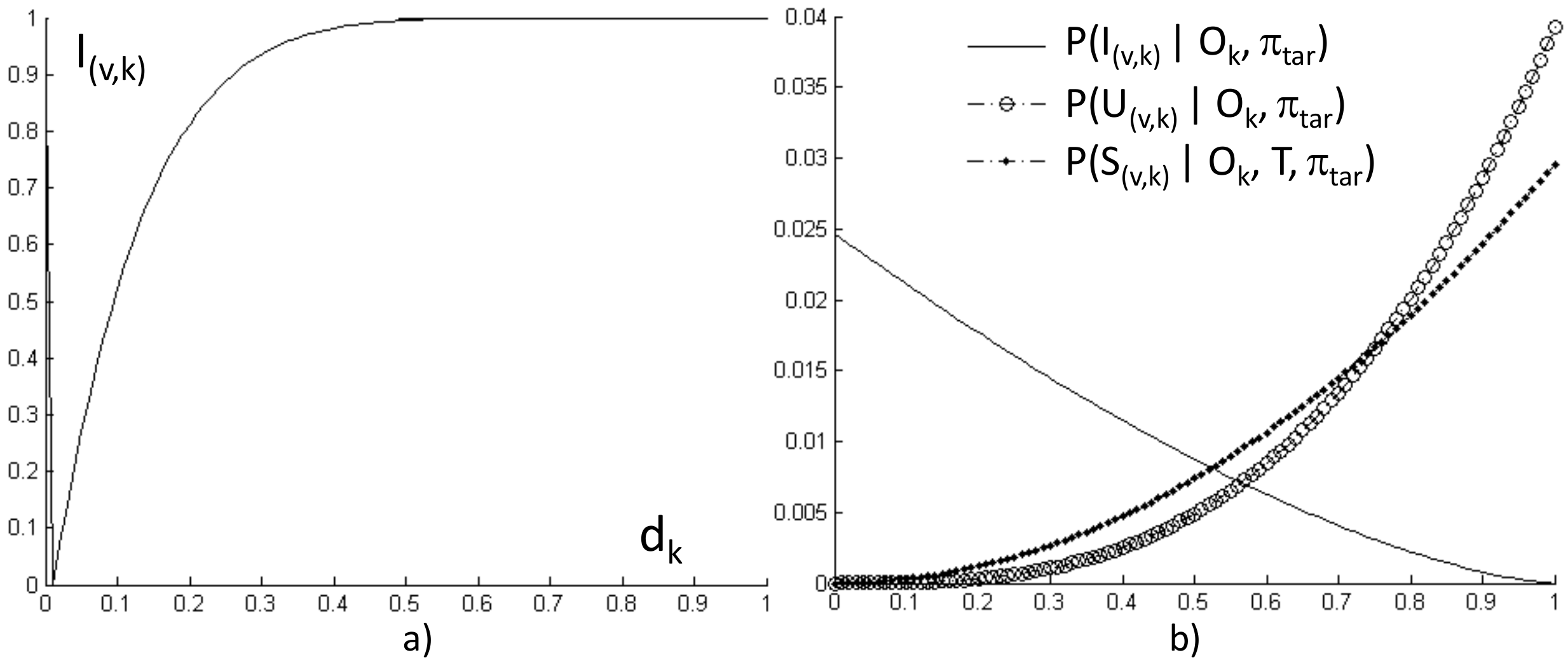}
  \caption{\textcolor{black}{a) Graphical representation of} $I_{(v,k)}$.  \textcolor{black}{b) Graphical representation of} $P(I_{(v,k)}|O_k,\pi_{tar})$, $P(U_{(v,k)}|O_k,\pi_{tar})$ and $P(S_{(v,k)}| O_k, T, \pi_{tar})$.}
  \label{fig:probability_distribution_functions_v0}
\end{figure}

As presented in figure  \ref{fig:Bayesian_Program_Network_Top_v0} b), $P(I_{(v,k)}|O_k,\pi_{tar})$ is described by a beta probability distribution function $\mathcal{B}_I$ characterized by the constants $\alpha_I=1$ and $\beta_I=2.5$.  The profile of the probability distribution function $P(I_{(v,k)}|O_k,\pi_{tar})$ is represented in figure \ref{fig:probability_distribution_functions_v0} b). The selected profile for $P(I_{(v,k)}|O_k,\pi_{tar})$ attributes higher probabilities for lower levels of $I_{(v,k)}$ and lower probabilities to higher values of $I_{(v,k)}$ in order to promote the selection of regions of the workspace with low values of inhibition level.

Following an analogous approach, $P(U_{(v,k)}|O_k,\pi_{tar})$ is described by a beta probability distribution function $\mathcal{B}_U$ (figure \ref{fig:probability_distribution_functions_v0} b)) with the constant parameters $\alpha_U=4$ and $\beta_U=1$. $P(U_{(v,k)}|O_k,\pi_{tar})$ attributes higher probability values to regions of the workspace perceived with higher uncertainty $U_{(v,k)}$.

$P(S_{(v,k)}| O_k, T, \pi_{tar})$  is described by a beta probability distribution function $\mathcal{B}_S$ defined by $\alpha_S=3$ and $\beta_S=1$ (figure \ref{fig:probability_distribution_functions_v0} b)), assigning higher probability values to workspace regions $v$ with higher values of saliency $S_{(v,k)}$.

\section{Experimental results}
\label{sec:experimental_results}

\subsection{Evaluation of the haptic stimulus perception model}
\label{sec:evaluation_of_the_haptic_stimulus}

As referred previously, this work extracts the parameters $\mu_E(M)$, $\sigma_E(M)$, $\mu_C(M)$, $\sigma_C(M)$ from the work \cite{Xu2013}. However, the Bayesian program proposed in this work to categorize the haptic stimulus $Material_1,\ldots, Material_n$ follows a different approach than the work \cite{Xu2013}.

Following an approach analogous to several previous works (Eg: \cite{Xu2013}, \cite{Bologna2013}, \cite{Liu2012}, \cite{Pape2012}) , the performance of the Bayesian model proposed in section \ref{ch:Bayesian_Program_I} and the consistency of the extracted parameters used to categorize the haptic stimulus  was evaluated by performing a numerical simulation of 400 trials consisting in the local haptic exploration of samples of each of the reference materials $Material_1,\ldots,Material_{10}$.  In each trial, the local haptic exploration of the reference materials was simulated by generating random samples $e^{'}_{(v, k)}$ and $c^{'}_{(v, k)}$, obtained from $e_{(v, k)}$ and $c_{(v, k)}$ corrupted with additive white Gaussian noise ($q_{C}$ and $q_{E}$), according the formulations $c^{'}_{(v, k)} = c_{(v, k)}+ q_{C}$ and $e^{'}_{(vi, k)}= e_{(v, k)} + q_{E}$, respectively.

As presented previously, $C_{(v, k)} \sim \mathcal{N}(\mu_{C}(M),\sigma_{C}(M))$ and $E_{(v, k)} \sim \mathcal{N}(\mu_{E}(M),\sigma_{E}(M))$ and in this experimental setup the additive white Gaussian noise is described by $Q_{C} \sim \mathcal{N}(0,\frac{\mu_{C}(M)}{2})$ and $Q_{E} \sim \mathcal{N}(0,\frac{\mu_{E}(M)}{2})$. For each reference material $Material_i$, the classification performance was evaluate for a initial exploration of that region $v$ $(k=0)$ and for progressive exploration of that region of the workspace $v$ during $(k=1,\ldots,4)$ exploration iterations. The categorization $\hat{m}_{(v, k)})$ of the perceived haptic stimulus is determined by \textit{MAP - Maximum a Posteriori} following the equation \ref{eq:categorizationHapticStimulus}.

\begin{align}
\hat{m}_{(v, k)} =\operatorname*{arg\,max}_{m_{(v, k)})} P(M_{(v, k)}| e^{'}_{(v, k)}, c^{'}_{(v, k)},\pi_{per})
\label{eq:categorizationHapticStimulus}
\end{align}

The evaluation of the performance of the Bayesian model $\pi_{per}$ proposed in section \ref{ch:Bayesian_Program_I} is presented in the confusion table  \ref{tb:bayesian_program_I_k1} using one exploration iterations ($k=0$) in that location $v$ and using five exploration iterations ($k=4$).

\begin{table*}
\scriptsize
\centering
\caption{Confusion table for the categorization of $Mat._i$ (ground truth) as $M.i$ (perceived category) by the Bayesian model $\pi_{per}$, using only one exploration sample $k=0$ (400 trials) and using five exploration samples $k=4$ (400 trials).}
\begin{tabular}{|l|c|c|c|c|c|c|c|c|c|c|c|c|c|c|c|c|c|c|c|c|}
\hline
&\multicolumn{2}{|c|}{\textit{M.1}}&\multicolumn{2}{|c|}{\textit{M.2}}&\multicolumn{2}{|c|}{\textit{M.3}}&\multicolumn{2}{|c|}{\textit{M.4}}&\multicolumn{2}{|c|}{\textit{M.5}}&\multicolumn{2}{|c|}{\textit{M.6}}&\multicolumn{2}{|c|}{\textit{M.7}}&\multicolumn{2}{|c|}{\textit{M.8}}&\multicolumn{2}{|c|}{\textit{M.9}}&\multicolumn{2}{|c|}{\textit{M.10}} \\ \hline
$k$&0&4&0&4&0&4&0&4&0&4&0&4&0&4&0&4&0&4&0&4\\ \hline \hline 
$Mat._1$&243&331 &0&1 &60&50 &0&0 &0&0 &0&0 &0&0 &0&0 &77&18 &20&0 \\ \hline \hline 
$Mat._2$&17&0 &176&316 &20&2 &0&0 &0&0 &45&0 &1&0 &4&0 &22&0 &115&82 \\ \hline \hline 
$Mat._3$&108&34 &2&0 &146&337 &0&0 &0&0 &0&0 &0&0 &0&0 &97&25 &47&4\\ \hline \hline 
$Mat._4$&0&0 &0&0 &0&0 &371&399 &0&0 &25&1 &4&0 &0&0 &0&0 &0&0\\ \hline \hline 
$Mat._5$&0&0 &0&0 &0&0 &0&0 &397&400 &0&0 &3&0 &0&0 &0&0 &0&0\\ \hline  \hline 
$Mat._6$&0&0 &11&0 &1&0 &24&0 &0&0 &257&397 &38&1 &29&0 &14&0 &26&2\\ \hline  \hline 
$Mat._7$&0&0 &1&0 &0&0 &6&0 &1&0 &19&0 &340&400 &32&0 &1&0 &0&0 \\ \hline \hline 
$Mat._8$&0&0 &0&0 &0&0 &0&0 &0&0 &10&0 &9&0 &381&400 &0&0 &0&0\\ \hline  \hline 
$Mat._9$&78&17 &1&1 &9&19 &0&0 &0&0 &0&0 &0&0 &0&0 &270&362 &42&1 \\ \hline \hline 
$Mat._{10}$&15&0 &65&92 &32&2 &0&0 &0&0 &42&1 &1&0 &6&0 &38&0 &201&305\\ \hline    
\end{tabular}
\label{tb:bayesian_program_I_k1}
\end{table*}

The results presented in table \ref{tb:bayesian_program_I_k1} show that globally the Bayesian model $\pi_{per}$ has a good capability to discriminate and categorize the perceived haptic stimulus with the correct category of reference materials $Material_i$. The Bayesian  model $\pi_{per}$ shows a worst classification performance for haptic stimulus $Material_2$, $Material_3$ and $Material_{10}$. By integrating a higher number of sensory samples $(k=4)$,  the global performance of the Bayesian model $\pi_{per}$ increases, including $Material_2$, $Material_3$ and $Material_{10}$ materials. The integration of  five sensory samples $(k=4)$ allows the system to improve the erroneous effect introduced by the uncertainties of the measurements and by the additive white Gaussian noise. As in other works by \cite{Xu2013}, \cite{Bologna2013}, \cite{Liu2012} and \cite{Pape2012}, the different materials are correctly discriminated with a high performance (average recognition rate higher than 90\%).

\begin{figure}[h]
  \centering
  \includegraphics[width=0.48\textwidth]{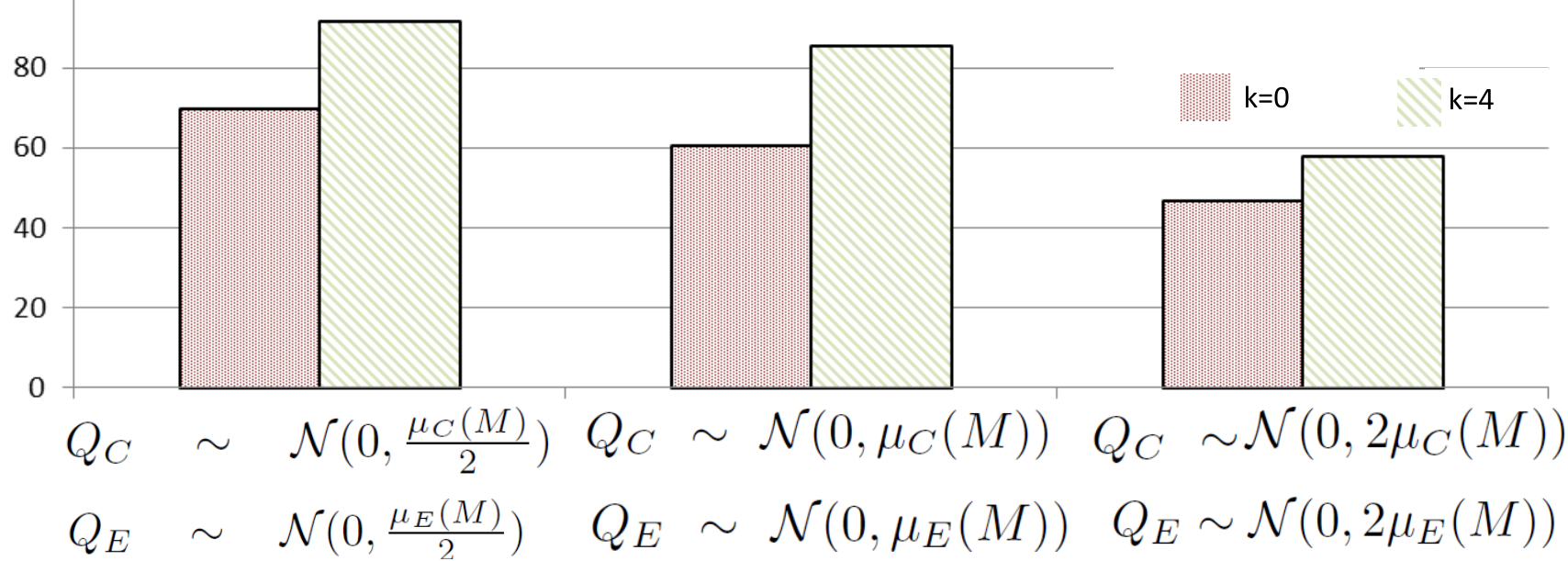}
  \caption{\textcolor{black}{Classification performance (average for 10 materials) of the Bayesian model} $\pi_{per}$\textcolor{black}{, using sensory samples corrupted with three different levels of additive white noise. The performance is evaluated integrating 1} ($k=0$) and 5 ($k=4$) \textcolor{black}{successive sensory samples.}}
  \label{fig:bars_classification_performance_v0}
\end{figure}

\textcolor{black}{In this work, we also have studied how the classification performance of the Bayesian model} $\pi_{per}$ \textcolor{black}{can be affected by increasing levels of additive white noise} $Q_{C}$ and $Q_{E}$. \textcolor{black}{The increasing levels of additive white Gaussian noise were simulated by increasing the standard deviation of the distributions of}  $Q_{C}$ and $Q_{E}$, \textcolor{black}{as presented in figure \ref{fig:bars_classification_performance_v0}. By increasing the magnitude of the standard deviation of the distributions of} $Q_{C}$ and $Q_{E}$ \textcolor{black}{the classification performance of the Bayesian model} $\pi_{per}$ \textcolor{black}{decreases. This effect is attenuated by the consecutive integration of several sensory samples ($k=4$). This demonstratates the relevancy of implementing an active haptic exloration strategy in order to promote the exploration of uncertainty regions of the workspace to improve the current perceptual representation.}

\subsection{Autonomous exploration of the workspace}
\label{Autonomous_Exploration}

\subsubsection{Visualization tools}
\label{sec:simulation_tools}
\begin{figure}[h]
  \centering
  \includegraphics[width=0.48\textwidth]{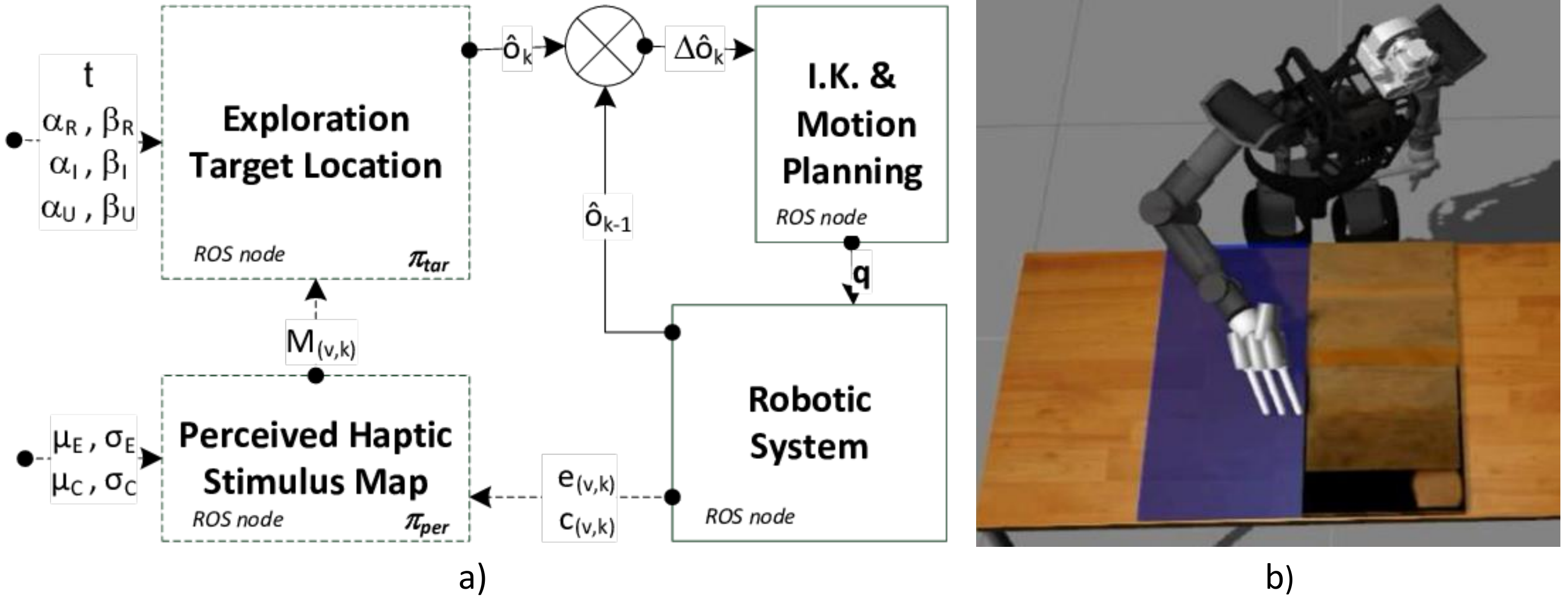}
  \caption{\textcolor{black}{a) High level control architecture (solid lines). The modules and variables involved in the determination of the reference signal} $\hat{o}_k$ \textcolor{black}{are represented with dashed line. b) Typical exploration behaviour of \textit{ATLAS} during the execution of the \textit{Scenario I} task} ($Trial_1$) \textcolor{black}{at time iteration} $k=0$.}
  \label{fig:control_and_exploration}
\end{figure}

The scientific concepts presented and described in this work have been tested in a virtual environment. The visualization tool selected to implement the virtual environment was \textit{Gazebo 1.7}. The robotic platform used in this work is the \textit{ATLAS} robotic platform \cite{atlas}, which is provided by the \textit{DARPA robotics challenge} software package \textit{DRCsim-2.5}. The high level control architecture of the system, that was implemented using \textit{Robotic Operating System-ROS Fuerte}, is presented in figure \ref{fig:control_and_exploration} a). The determination of the inverse kinematics (IK), motion planing and lower level control of the \textit{ATLAS} robotic platform are not discussed in this work.

\subsubsection{Haptic stimulus scenarios}
\label{sec:static_stimulus}
The virtual environment built for this work consists in the robotic manipulation platform \textit{ATLAS} and a haptic stimulus presented in the top of a planar table placed in front of the robotic platform. The workspace region is partitioned in a volumetric grid as suggested in figure \ref{fig:workspace_general_grid}. In this work, the workspace volumetric grid has dimensions ${X}^W_{l}=0m$, ${X}^W_{u}=0.30m$, ${Y}^W_{l}=0m$,${Y}^W_{u}=0.60m$ , ${Z}^W_{l}=0m$, ${Z}^W_{u}=0.01m$ (figure \ref{fig:control_and_exploration} b) ). Each voxel (cube) has a side dimension of $\varepsilon=0.01m$.

\begin{figure*}
  \centering
  \includegraphics[width=0.75\textwidth]{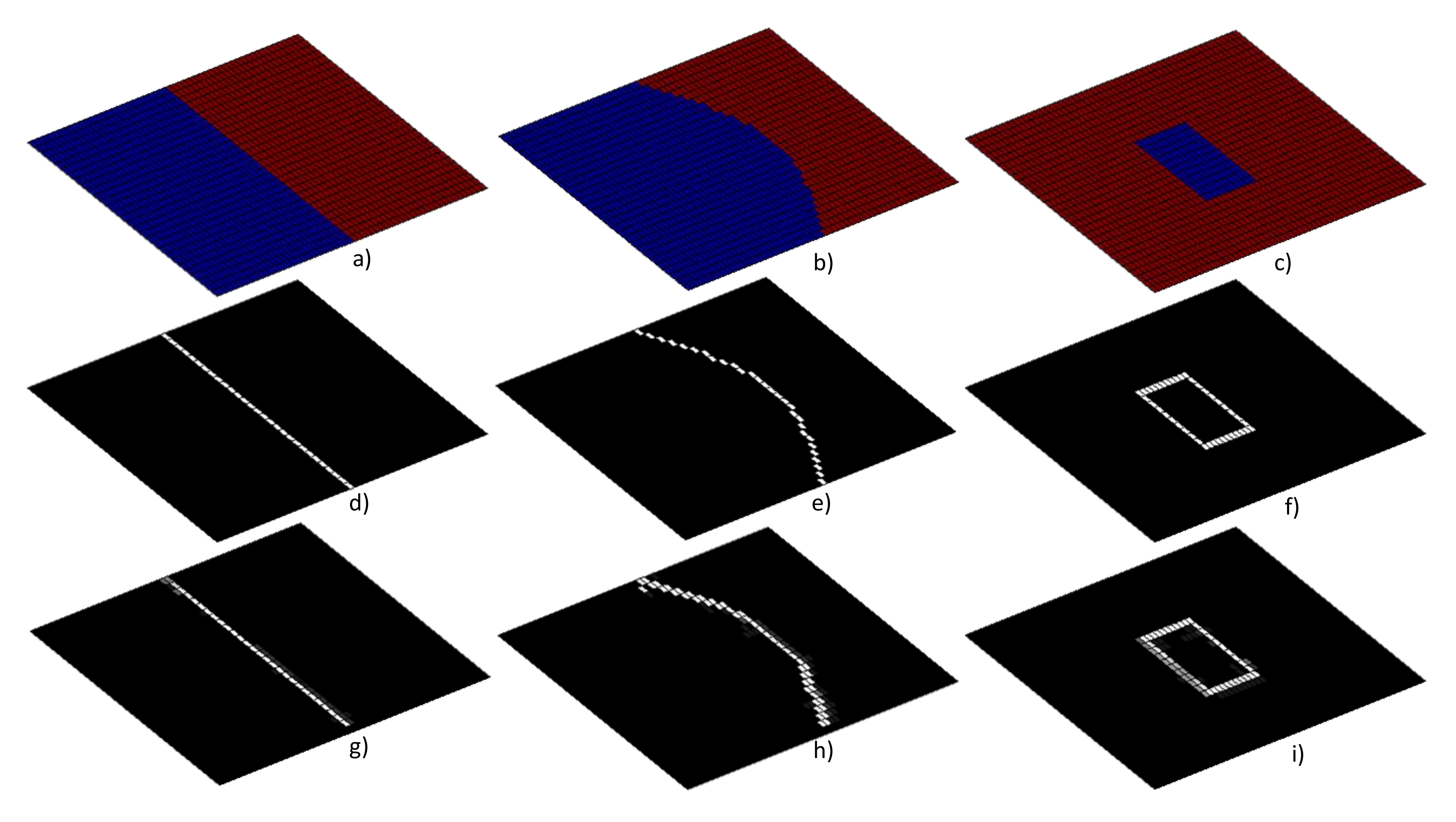}
  \caption{a)-c) Representation of the haptic stimulus proposed in \textit{Scenario I}, \textit{Scenario II} and \textit{Scenario III}. d)-f) Benchmark exploration paths, $\textbf{U}$ for $T=t=$ \textit{"Search and follow of discontinuities between regions of surfaces with $Material_8$ (silicone) and $Material_{10}$ (wood)"} in \textit{Scenario I}, \textit{II} and \textit{III}. g)-i) Exploration paths performed during 10 experimental trials for $T=t=$ \textit{"Search and follow of discontinuities between regions of surfaces with $Material_8$ and $Material_{10}$"} in \textit{Scenario I}, \textit{II} and \textit{III}. Dark colors represent regions visited few times. Light colors represent regions visited many times.}
  \label{fig:workspace_stimulus_ground_results}
\end{figure*}

The surface presented in the workspace region is made of two distinct materials: $Material_8$ (blue silicone) and $Material_{10}$ (wood). Three different configurations (\textit{Scenario I}, \textit{Scenario II} and \textit{Scenario III}) of the haptic stimulus are presented in figure \ref{fig:workspace_stimulus_ground_results} a)-c).

\subsubsection{Autonomous exploration performance}
\label{sec:performance}
This work assumes that, at each time iteration step $k$, a exploratory element of the robotic right hand touches a workspace region $v$, sensory samples $e_{(v,k)}$ and $c_{(v,k)}$ are artificially synthesised  from the respective probability distribution functions $P(E_{(v, k)}|m_{(v, k)},\pi_{per})$ and $P(C_{(v, k)}|m_{(v, k)},\pi_{per})$, given the known ground truth material $m_{(v, k)}$ for that region of the workspace. In each of the scenarios the index fingertip of the \textit{ATLAS} robotic platform is initialized ($k=0$) at different locations. In \textit{Scenario I}, $\textbf{v}_0=(28,30,0)$, in \textit{Scenario II}, $\textbf{v}_0=(28,31,0)$ and in \textit{Scenario I}, $\textbf{v}_0=(17,35,0)$.  The probability distribution function $P(O_k | t, S_{(v,k)},I_{(v,k)}, U_{(v,k)}, \pi_{tar})$ is initialized as uniform probability distribution function.

The exploration paths used as benchmarks , $\mathcal{B}$, for each of the scenarios (\textit{Scenario I}, \textit{Scenario II} and \textit{Scenario III}) are represented in the figure   \ref{fig:workspace_stimulus_ground_results} d)-f) respectively, corresponding to the edge regions between $Material_8$  and $Material_{10}$.

The different configurations of stimulus (unknown \textit{a priori} to the robotic system) have been explored during 10 different trials, for each of the scenarios \textit{Scenario I}, \textit{II} and \textit{III}. The error for each exploration trial has been evaluated using a performance metric , $\Gamma$, proposed in equation \ref{eq:objective_definition}. For each trial, the exploration procedures lasts $l$ time iterations ($k=0,\ldots, (l-1)$). In order to compare the performance of the exploration in the different scenarios, a normalized performance metric has been determined, $\Gamma \setminus l$. This metric represents the average divergence in $cm$ per time iteration of the exploratory element relatively to the ground truth.

Following an experimental approach analogous to \cite{Li2013}, the results of these experimental sessions are compiled and presented in table \ref{tb:exploration_results_scenarios}. Table \ref{tb:exploration_results_scenarios} shows that the proposed Bayesian model $\pi_{tar}$ has a good precision and simultaneously a considerable generalization capability. The exploration task was performed with an average divergence ($\Gamma \setminus l$) from the ground truth exploration paths smaller than $1 cm$. The scenarios \textit{Scenario I} and \textit{Scenario III} have lower divergence values due to the lower number of slope variation regions in the discontinuity between the two regions.

\begin{table}[h]
\scriptsize
\centering
\caption{Performance of the exploration procedures performed  in \textit{Scenarios I}, \textit{II} and \textit{III}.}
\begin{tabular}{|l|c|c|c|c|c|c|c|c|c|c|}
\hline
&\multicolumn{3}{|c|}{\textbf{Scenario I}}&\multicolumn{3}{|c|}{\textbf{Scenario II}}&\multicolumn{3}{|c|}{\textbf{Scenario III}} \\
\hline
\textbf{\textit{Trial}}& $l$ & $\Gamma$ & $\Gamma/l$ & \textit{l}& $\Gamma$ & $\Gamma/l$& \textit{l}& $\Gamma$& $\Gamma/l$\\ \hline 
 \textbf{1}&35&7.0&0.2&50&29.1&0.6&43&4.0&0.1\\ \hline 
\textbf{2}&37&9.0&0.2&50&24.8&0.5&49&21.0&0.4\\ \hline 
\textbf{3}&40&17.0&0.4&46&22.8&0.5&42&2.0&0.0\\ \hline 
\textbf{4}&29&1.0&0.0&46&22.8&0.5&51&13.0&0.3\\ \hline 
\textbf{5}&27&1.0&0.0&53&34.9&0.6&46&9.0&0.2\\ \hline 
\textbf{6}&29&1.0&0.0&46&22.8&0.5&48&14.0&0.3\\ \hline 
\textbf{7}&30&2.0&0.0&48&25.1&0.5&50&12.0&0.2\\ \hline 
\textbf{8}&28&1.0&0.4&62&43.1&0.7&47&11.0&0.2\\ \hline 
\textbf{9}&29&2.0&0.1&54&37.7&0.7&43&3.0&0.1\\ \hline 
\textbf{10}&28&1.0&0.0&47&23.8&0.5&53&14.0&0.3\\ \hline 
\textbf{$\mu$}&31.2&4.2&0.2&50.2&28.6&0.6&47.2&10.3&0.2\\ \hline  
\textbf{$\sigma$}&4.5&5.3&0.1&5.0&7.2&0.0&3.7&5.9&0.1\\ \hline    
\end{tabular}
\label{tb:exploration_results_scenarios}
\end{table}

By performing a empirical comparison between figure \ref{fig:workspace_stimulus_ground_results} g)-i) and figure \ref{fig:workspace_stimulus_ground_results} d)-f), we can verify that the experimental  exploration paths have a very good structural correspondence with the benchmark exploration paths in all of the 3 scenarios.  In \textit{Scenario III},  the robotic system tracks the complete structure (loop closure) of the discontinuity (closed curve). As in the work \cite{Martinez2013}, the system was able to deal with severe changes in the slop of the discontinuity. In \textit{Scenario II} the robotic system was able to track a haptic discontinuity with a progressive inversion in the slop of the discontinuity, what clearly demonstrates the generalization capability of the proposed approach. This emergent behaviour of this system presents an improvement of the results presented in \cite{Martinez2013}. The test of the system with other slop variations in discontinuities than right angles ($90^\circ$) was suggested by \cite{Martinez2013} as future work.

Videos showing the detailed representation and temporal evolution of  $P(I_{(v,k)}|O_k,\pi_{tar})$, $P(S_{(v,k)} | O_k, T,\pi_{tar})$ and $P(U_{(v,k)} | O_k, \pi_{tar})$ involved in the determination of $P(O_k | t, s_{(v,k)},i_{(v,k)}, u_{(v,k)}, \pi_{tar})$ and inference of $\hat{o}_k$,  are available online \textit{www.rmartins.net/iros2014a}.

\section{Conclusions and future work}
\label{sec:conclusions_futurework}
This work has presented the theoretical foundations and experimental implementation of the Bayesian models of the touch attention mechanisms involved in the active haptic exploration of heterogeneous surfaces by generic robotic hands and sensory apparatus. 

The global architecture of the proposed models have shown a good generalization capability during the execution of a discontinuity following task between surface regions made of distinct materials with three different spatial configurations. The system was also able to perceive and discriminate 10 different classes of materials. The system have shown the capability to overcome the challenges placed by the uncertainty associated to the unknown structure of the environment and noisy sensory signals. 

In the next developments of this work, a new Bayesian model will be introduced in the action-perception loop architecture of the system presented in figure \ref{fig:Architecture_and_information_flow_v1_5}. The new model will be related with the recognition of the identity of the explored structure (using the shape of discontinuities as main cue) during the haptic exploration of the surface: active haptic exploration and active recognition of objects.

\bibliographystyle{IEEEtran}
\bibliography{bibRicardoMartins_v15}

\end{document}